\documentclass[conference]{IEEEtran}

\usepackage{amsmath,graphicx,epstopdf}
\usepackage{subfigure}
\usepackage{color}
\usepackage{overpic}
\usepackage[table,xcdraw]{xcolor}
\usepackage{multirow}
\usepackage{overpic}
\usepackage{booktabs}
\usepackage{url}
\usepackage{array}

\IEEEoverridecommandlockouts
\begin{document}
%
\title{Dual Recovery Network with Online Compensation \\ for Image Super-Resolution}

\author{\IEEEauthorblockN{Sifeng Xia$^1$, Wenhan Yang$^1$, Jiaying Liu$^{1,\ast}$ and Zongming Guo$^{1,2}$
\thanks{\footnotesize{$^{\ast}$Corresponding author \newline
This work was supported by National Natural Science Foundation of China under contract No.U1636206. We also gratefully acknowledge the support of NVIDIA Corporation with the GPU for this research.}}
}
\IEEEauthorblockA{$^1$Institute of Computer Science and Technology, Peking University, Beijing, China\\
$^2$Cooperative Medianet Innovation Center, Shanghai, China}}

\maketitle


\begin{abstract}
Image super-resolution (SR) methods essentially lead to a loss of some high-frequency (HF) information when predicting high-resolution (HR) images from low-resolution (LR) images without using external references. To address this issue, we additionally utilize online retrieved data to facilitate image SR in a unified deep framework. A novel dual high-frequency recovery network (DHN) is proposed to predict an HR image with three parts: an LR image, an internal inferred HF (IHF) map (HF missing part inferred solely from the LR image) and an external extracted HF (EHF) map. In particular, we infer the HF information based on both the LR image and similar HR references which are retrieved online. For the EHF map, we align the references with affine transformation and then in the aligned references, part of HF signals are extracted by the proposed DHN to compensate for the HF loss. Extensive experimental results demonstrate that our DHN achieves notably better performance than state-of-the-art SR methods.
\end{abstract}

%
\IEEEpeerreviewmaketitle

\begin{figure*}[t]
\centering
\includegraphics[width=160mm]{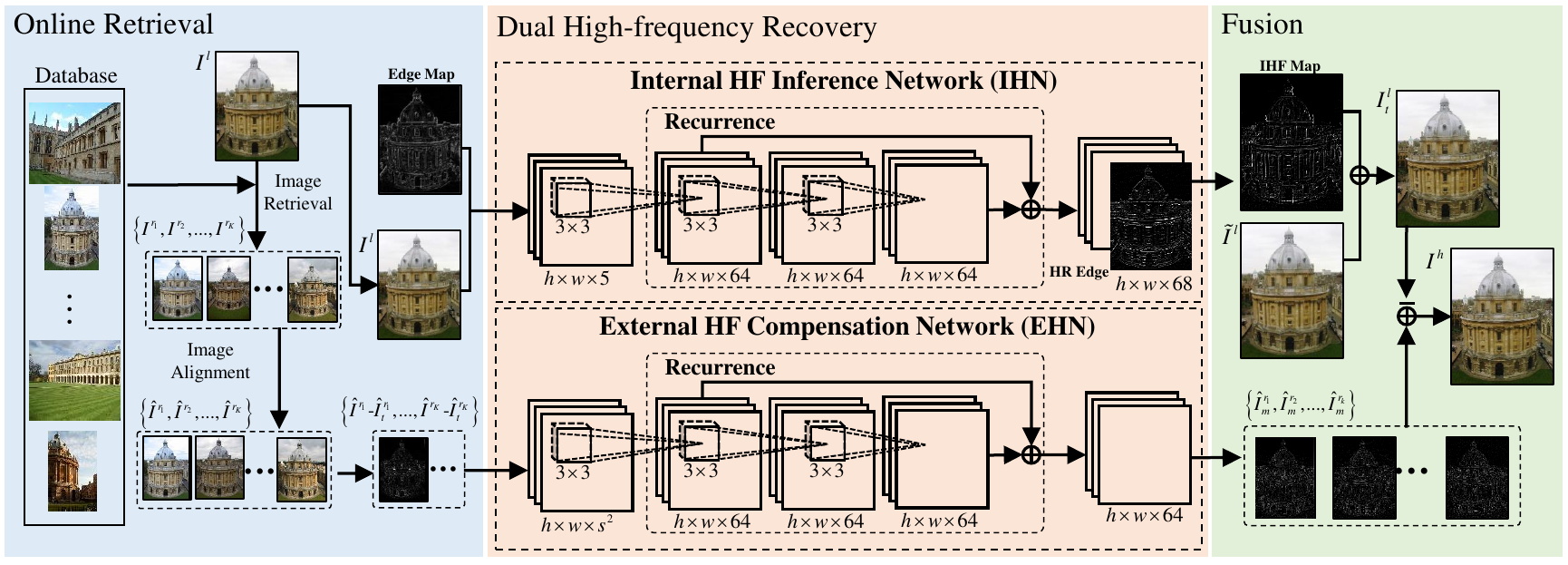}
\caption{Framework of the proposed SR method based on the dual high-frequency recovery network (DHN) with online compensation. ${h}\times{w}\times{\ast}$ means size of the convolution layer and $\ast$ represents channel numbers. $s$ is the magnification factor. Image $\tilde{I}^{l}$ is simply up-sampled form $I^{l}$. $I^{l}_{t}$ is the intermediate image derived by IHN and $I^{h}$ is the further enhanced result image.}
\label{fig:f1}
\vspace{-4mm}
\end{figure*}

\section{Introduction}
Image super-resolution (SR) aims to estimate a high-resolution (HR) image from low-resolution (LR) observations. In essence, due to the information loss in the image degradation process, SR is an ill-posed problem. The earliest works, image interpolation, estimate the HR image based on local statistics of the LR image. Typical methods include bilinear, bicubic and new edge directed interpolation that predict the HR pixels by utilizing the spatial relationship between LR and HR pixels. Later on, many successive works \cite{sun_gradient_2011,Breg2} regard the image SR as a Maximum-a-posteriori estimation and propose to impose various priors to constrain the inverse estimation of image SR. In these methods, priors and constraints are typically achieved in a heuristic way. Thus, it is insufficient to represent the diversified patterns of natural images.

Learning based methods obtain a mapping between LR and HR images based on a large training set with dynamic learned prior knowledge. Sparse representation based methods such as \cite{SCSR} learn the map by building an LR and HR patch mapping dictionary. Neighbor embedding (NE) methods linearly combine the HR neighbors to infer the HR image. Timofte \textit{et al.} \cite{aplus} proposed an adjusted anchored neighborhood regression method for image SR. Li \textit{et al.} \cite{NE} proposed a neighbor preserving based method which specially utilizes HR reference patches only in reconstructing the high frequency region of LR images. Recently, deep-learning based methods \cite{SRCNN,DSP,VDSR,IHN,CVIUVDSR} are proposed. SRCNN is the first method \cite{SRCNN} that utilizes a three-layer convolutional network for image SR. In \cite{DSP}, the sparse prior is incorporated into the network. Then, the residual learning \cite{VDSR} and sub-band recovery with edge guidance \cite{IHN} networks are constructed to recover HF signal and offer state-of-the-art performance.

Despite impressive results achieved by the learning-based methods, some HF information has still been lost because of the ill-posed nature of the image SR and the problem that mean squared error leads to \emph{regression to mean} \cite{timofte2016semantic}. As a result, a few methods have recently been proposed, which additionally compensate for HF information loss with online retrieved HR references. Yue \textit{et al.} \cite{landmark2013} directly utilized the references to enhance the SR result by patch matching and patch blending. Li \textit{et al.} \cite{SRreference} used the retrieved HR image patches to learn more accurate sparse distribution. Liu \textit{et al.} \cite{onlinetmm2016} utilized a group-structured sparse representation to further use the nonlocal dependency information of HR references. However, in these methods there are still several important issues not being fully considered. For example, their fusion methods do not effectively extract external HF information for compensation, which may even bring artifacts. Besides, they did not make full use of the internal redundancy to benefit the recovery of HF information.

To address the aforementioned issues, we propose a unified deep network that additionally utilizes online retrieved data to facilitate image SR. Our work can efficiently extract an HF map from multiple HR references that are retrieved based on the intermediately inferred SR image.

Contributions of this paper are as follows: 1) It is the first work that efficiently extracts high-frequency information from the HR reference and successfully compensate for the HF information loss of the SR result with the deep framework. 2) We show the proposed method is capable to model internal and external images jointly, achieving a more accurate and robust fusion of internal and external information for HF information recovery. 3) Compared with both previous deep learning-based methods and online compensation SR methods, our approach has offered new state-of-the-art performance.

The rest of the article is organized as follows. Sec. \ref{sec:DHN} illustrates our DHN network. Details of utilizing the EHF map for compensation are introduced in Sec. \ref{sec:online}. Experimental results are shown in Sec. \ref{sec:experiment} and concluding remarks are given in Sec. \ref{sec:conclu}.

\section{Dual High-Frequency Recovery Network}
\label{sec:DHN}

Given an LR image $I^{l}$, we predict the HR image $I^{h}$ from $I^{l}$ with the reference of $\emph{K}$ retrieved HR reference images $\{I^{r_{1}},I^{r_{2}},...,I^{r_{\emph{K}}}\}$ by our dual high-frequency recovery network (DHN). Architecture of the proposed DHN has been illustrated in Fig. \ref{fig:f1}. DHN consists of two components called internal high-frequency inference network (IHN) and external high-frequency compensation network (EHN), respectively. IHN infers missing HF information of $I^{l}$ merely based on internal data in $I^{l}$. Then, the intermediate SR image $I_{t}^{l}$ is generated by combining the internal inferred HF (IHF) map and the simply up-smapled LR image $\tilde{I}^{l}$. EHN further enhances the final SR result $I^{h}$ by adding the external extracted HF (EHF) map obtained from the aligned retrieved HR reference images $\{\hat{I}^{r_{1}},\hat{I}^{r_{2}},...,\hat{I}^{r_{\emph{K}}}\}$ to the intermediate image $I_{t}^{l}$.


\subsection{Internal High-Frequency Inference Network}
\label{sec:IHN}
The first component IHN proposed by \cite{IHN} is utilized to initially reconstruct the LR image $I^{l}$ with its own information. As shown in Fig.1, $I^{l}$ and its edge map, which is extracted by applying a hand-crafted edge detector, are utilized as the input of IHN. Then, the recurrent network of IHN estimates the IHF map from the above input. IHN also predicts an HR edge map, which is used to further guide the HF map estimation.

With the inferred IHF map, the intermediate result image $I_{t}^{l}$ is then generalized as follows:
\begin{equation}
\label{eq1}
I_{t}^{l} = \tilde{I}^{l} \oplus \varphi(I^{l}),
\end{equation}
where $\oplus$ is the sum operation and $\varphi(I^{l})$ represents the process that IHN infers IHF map from LR image $I^{l}$. $\tilde{I}^{l}$ is the image that simply up-sampled from $I^{l}$. We then define the loss of IHN as the combination of loss of the predicted HR edge and $I_{t}^{l}$. The loss is measured by the mean squared error (MSE) with the ground truth signal.

\subsection{External High-Frequency Compensation Network}
\label{sec:EHN}

IHN works well in predicting the HF map from an LR image. However, during this process not all HF information can be well recovered. This inspires us to construct EHN to further extract the significant EHF map $\hat{I}^{r}_{m}$ from each HR reference $\hat{I}^{r}$. Note that during training process $\hat{I}^{r}$ is generated from the ground truth HR image.

It's common for an LR image and its reference image to have illumination and color differences. Moreover, there is much useless low-frequency information in the reference that may affect HF information extraction. Therefore we take different measures to improve the robustness of the process of extracting $\hat{I}^{r}_{m}$. First, contrast of the label images is additionally adjusted to simulate the common illumination and color differences in training process. Besides, we alternatively utilize the difference image between $\hat{I}^{r}$ and its intermediate SR image $\hat{I}^{r}_{t}$ as the input of EHN, rather than directly input the information of $\hat{I}^{r}$. $\hat{I}^{r}_{t}$ is obtained through up-sampling the down-sampled image of $\hat{I}^{r}$ by IHN. The difference image is chosen because of its high efficiency in reducing illumination and color differences and removing redundant low-frequency information.

Then, EHN extracts the EHF map from the input by the recurrent network. Final reconstructed result $I^{h}$ is derived by:
\begin{equation}
\label{eq2}
I^{h} = I_{t}^{l} \bar{\oplus} \psi(\hat{I}^{r}-\hat{I}^{r}_{t}),
\end{equation}
where $\psi$ is the formulation of the process that EHN extracts the HF map $\hat{I}^{r}_{m}$. The operation $\bar{\oplus}$ represents the combination of the intermediate image $I_{t}^{l}$ and $\hat{I}^{r}_{m}$. During the training process, $\hat{I}^{r}_{m}$ is directly added to $I_{t}^{l}$. In the testing process, $\hat{I}^{r}_{m}$ is utilized based on patch matching results, which is elaborated in Sec. \ref{sec:utilization}. Loss of EHN is defined as MSE between $I^{h}$ and the raw ground truth image.

\section{Online Compensation}
\label{sec:online}
Different with the training process, we retrieve HR reference images $\{I^{r_{1}},I^{r_{2}},...,I^{r_{\emph{K}}}\}$ online for compensation with the method proposed in \cite{onlinetmm2016} during the testing process. Then, the aligned HR references $\{\hat{I}^{r_{1}},\hat{I}^{r_{2}},...,\hat{I}^{r_{\emph{K}}}\}$ are derived by aligning each $I^{r}$ to $I_{t}^{l}$ and the HF maps $\{\hat{I}^{r_{1}}_{m},\hat{I}^{r_{2}}_{m},...,\hat{I}^{r_{\emph{K}}}_{m}\}$ are later extracted from the aligned references. As pixels in each aligned reference $\hat{I}^{r}$ are still not exactly corresponding to the pixels at the same position of $I_{t}^{l}$ ,extracted feature values of $\hat{I}^{r}_{m}$ can not be directly added to the intermediate up-sampled image $I_{t}^{l}$. Thus patch matching is used to guide the combination of $\hat{I}^{r}_{m}$ and $I_{t}^{l}$.

\subsection{Patch Matching}
\label{sec:pmatch}
There are usually significant differences on illumination, color and resolution between the intermediate SR image $I_{t}^{l}$ and each aligned HR reference $\hat{I}^{r}$. As a result, for the purpose of better matching results we first utilize the intermediate SR reference image $\hat{I}^{r}_{t}$ mentioned in Sec. \ref{sec:EHN} that shares similar resolution-level with $I_{t}^{l}$ for matching. Then, we adjust $\hat{I}^{r}_{t}$ to reduce the effect of illumination difference:
\begin{equation}
\label{eq3}
\hat{I}^{r'}_{t} = (\hat{I}^{r}_{t} - \tau(\hat{I}^{r}_{t}))\frac{\sigma(I_{t}^{l})}{\sigma(\hat{I}^{r}_{t})}+\tau(I_{t}^{l}),
\end{equation}
where $\hat{I}^{r'}_{t}$ is the transform result, $\tau(\cdot)$ and $\sigma(\cdot)$ are the mean and standard deviation values of all pixels of the image, respectively. Then, $I_{t}^{l}$ is split into overlapped query patches of size $\sqrt{n}\times\sqrt{n}$ at the step size $4$. And we search for the corresponding patches of the query patches within a search window in $\hat{I}^{r'}_{t}$.

Since small patches contain little structural information of raw images, patch matching results at small patch size are not accurate. Thus we perform patch matching between $I_{t}^{l}$ and $\hat{I}^{r'}_{t}$ with large patches. Considering it is impossible for each patch in $I_{t}^{l}$ to have an exact corresponding large patch in $\hat{I}^{r'}_{t}$, a method that adaptively adjusts patch sizes according to patch difference \cite{landmark2013} is adopted for more accurate patch matching.

Let $\mathbf{P}_{i}$ denote the query patch of size $\sqrt{n}\times\sqrt{n}$ in $I_{t}^{l}$ centered at position $i$ and $\mathbf{Q}_{j}^{i}$ denote the $\sqrt{n}\times\sqrt{n}$ candidate patch in $\hat{I}^{r'}_{t}$ centered at $j$. We search for the best matching candidate patch of $\mathbf{P}_{i}$ within the search window of size $3\sqrt{n}\times3\sqrt{n}$ centered at $i$ in $\hat{I}^{r'}_{t}$. The patch distance between $\mathbf{P}_{i}$ and $\mathbf{Q}_{j}^{i}$ is defined as:
\begin{equation}
\label{eq4}
d(\mathbf{P}_{i},\mathbf{Q}_{j}^{i}) = ||\mathbf{P}_{i}-\mathbf{Q}_{j}^{i}||^{2}_{2}+\rho||\nabla(\mathbf{P}_{i})-\nabla(\mathbf{Q}_{j}^{i})||^{2}_{2},
\end{equation}
where $\nabla$ is the operation that calculates the gradient of the patches and $\rho$ is the weighting parameter, which is set to be $10$ in this paper. Besides, DC components of the patches are removed before distance computation.

The value of $d(\mathbf{P}_{i},\mathbf{Q}_{j}^{i})/(\sqrt{n}\times\sqrt{n})$ is defined as gradient mean square error (GMSE) and $G_{i}^{min}$ is set as minimum GMSE value between the query patch $\mathbf{P}_{i}$ and the candidate patch $\mathbf{Q}_{j}^{i}$. Patch matching is performed at initial size $21\times21$ and changed to a smaller size if the value of $G_{i}^{min}$ is too large according to Eq. \ref{eq5}.
\begin{equation}
\label{eq5}
\sqrt{n}=\left\{
\begin{aligned}
& 21, & &G_{i}^{min} <= 200, \\
& 17, & &200 < G_{i}^{min} <= 500, \\
& 13, & &500 < G_{i}^{min} <= 800, \\
& 9,  &  &G_{i}^{min} > 800. \\
\end{aligned}
\right.
\end{equation}
 The sliding step of patch matching is set to be $\sqrt{n}/3$. Then, a closest candidate patch $\mathbf{Q}_{j_{0}}^{i}$ is found. However, a large step size may result in missing a better matching patch in $\hat{I}^{r'}_{t}$. Thus we further search a candidate patch of the same size as $\mathbf{Q}_{j_{0}}^{i}$ within a $(2\times\sqrt{n}/3-1)^{2}$ size search window centered at position $j_{0}$ in $\hat{I}^{r'}_{t}$, with the step size of $1$.

\subsection{External High-Frequency Information Utilization}
\label{sec:utilization}
After patch matching, pixels at the same position in the matched patches between $I_{t}^{l}$ and $I_{t}^{r'}$ are matched. Then, the EHF maps are combined with $I_{t}^{l}$ based on the pixel-wise matching correlation. For each pixel $\mathbf{p}$ in $I^{l}_{t}$, we define the set of its matching pixels in \emph{K} EHF maps as $\Omega_{\mathbf{p}}$. Then, the final fused external HF map $I^{l}_{m}$ that can be directly added to $I_{t}^{l}$ is obtained by:
\begin{equation}
\label{eq6}
I^{l}_{m,\mathbf{p}}=\left\{
\begin{aligned}
& \frac{\sum \limits_{\mathbf{q}\in\Omega_{\mathbf{p}}}\hat{I}^{r}_{m,\mathbf{q}} \cdot e^{\frac{-d(\mathbf{p},\mathbf{q})}{100}}}{\sum \limits_{\mathbf{q}\in\Omega_{\mathbf{p}}} e^{\frac{-d(\mathbf{p},\mathbf{q})}{100}}}, &|\Omega_{\mathbf{p}}|\neq0, \\
& 0, &|\Omega_{\mathbf{p}}|=0.
\end{aligned}
\right.
\end{equation}
$|\Omega_{\mathbf{p}}|$ represents the number of elements in set $\Omega_{\mathbf{p}}$. $d(\mathbf{p},\mathbf{q})$ is the GMSE value between the patches that $p$ and $q$ belong to.

Finally the result SR image is obtained by directly adding the final extracted HF map $I^{l}_{m}$ to the intermediate reconstructed SR image $I_{t}^{l}$ as $I^{h}={I_{t}^{l}}\oplus{I^{l}_{m}}$.

\begin{figure}[t]
\centering
\subfigure[]{
\includegraphics[height=16mm]{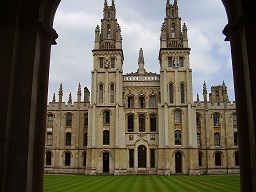}}
\subfigure[]{
\includegraphics[height=16mm]{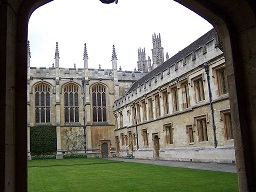}}
\subfigure[]{
\includegraphics[height=16mm]{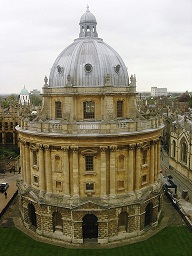}}
\subfigure[]{
\includegraphics[height=16mm]{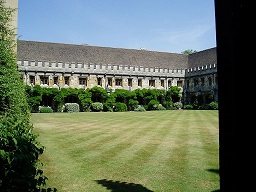}}
\subfigure[]{
\includegraphics[height=18mm]{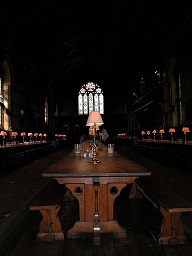}}
\subfigure[]{
\includegraphics[height=18mm]{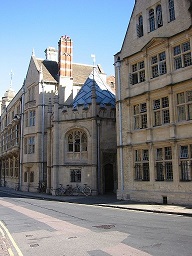}}
\subfigure[]{
\includegraphics[height=17mm]{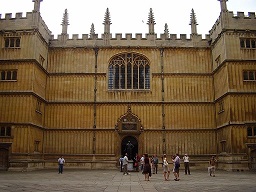}}
\subfigure[]{
\includegraphics[height=17mm]{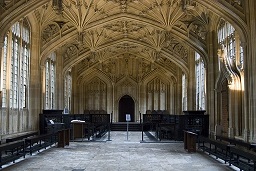}}
\caption{Testing images from (a) to (h).}
\label{fig:f3}
\vspace{-6mm}
\end{figure}

\vspace{2mm}
\section{Experimental Results}
\label{sec:experiment}

\begin{figure*}[t]
\scriptsize
\centering
\subfigure{
\begin{overpic}[height=27mm]{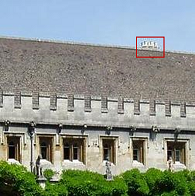}
\put(50,0){\includegraphics[width=14mm,clip]{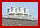}}
\end{overpic}
}
\subfigure{
\begin{overpic}[height=27mm]{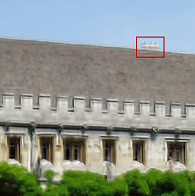}
\put(50,0){\includegraphics[width=14mm,clip]{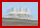}}
\end{overpic}
}
\subfigure{
\begin{overpic}[height=27mm]{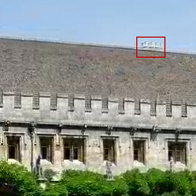}
\put(50,0){\includegraphics[width=14mm,clip]{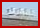}}
\end{overpic}
}
\subfigure{
\begin{overpic}[height=27mm]{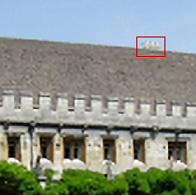}
\put(50,0){\includegraphics[width=14mm,clip]{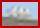}}
\end{overpic}
}
\subfigure{
\begin{overpic}[height=27mm]{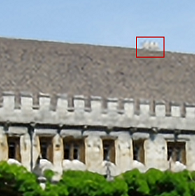}
\put(50,0){\includegraphics[width=14mm,clip]{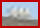}}
\end{overpic}
}
\subfigure{
\begin{overpic}[height=27mm]{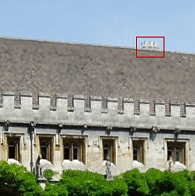}
\put(50,0){\includegraphics[width=14mm,clip]{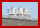}}
\end{overpic}
}\addtocounter{subfigure}{-6}
\subfigure[Ground Truth]{
\begin{overpic}[height=27mm]{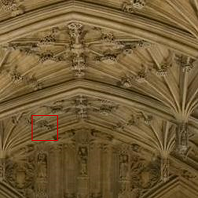}
\put(50,0){\includegraphics[width=14mm,clip]{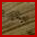}}
\end{overpic}
}
\subfigure[NE \cite{NE}]{
\begin{overpic}[height=27mm]{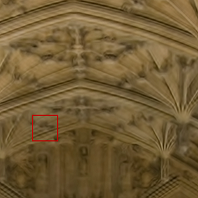}
\put(50,0){\includegraphics[width=14mm,clip]{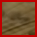}}
\end{overpic}
}
\subfigure[Landmark \cite{landmark2013}]{
\begin{overpic}[height=27mm]{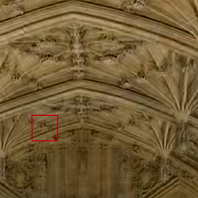}
\put(50,0){\includegraphics[width=14mm,clip]{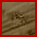}}
\end{overpic}
}
\subfigure[GSSR \cite{onlinetmm2016}]{
\begin{overpic}[height=27mm]{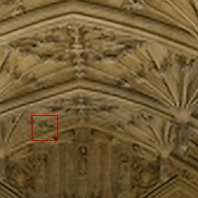}
\put(50,0){\includegraphics[width=14mm,clip]{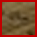}}
\end{overpic}
}
\subfigure[Baseline \cite{IHN}]{
\begin{overpic}[height=27mm]{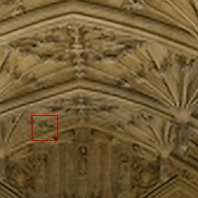}
\put(50,0){\includegraphics[width=14mm,clip]{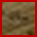}}
\end{overpic}
}
\subfigure[Proposed Method]{
\begin{overpic}[height=27mm]{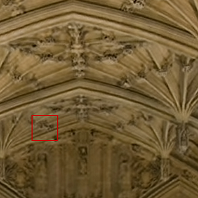}
\put(50,0){\includegraphics[width=14mm,clip]{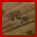}}
\end{overpic}
}
\caption{Subjective results of different methods with magnification factor 3 for testing images Fig. \ref{fig:f3}(d) and Fig. \ref{fig:f3}(h). Some regions that have HF signal have been marked in red rectangle and enlarged for comparison.}
\label{fig:f4}
\vspace{-4mm}
\end{figure*}

\subsection{Experimental Settings}
\label{sec:experimentset}
We train our DHN based on 91 images in \cite{SCSR} and 200 training images in \emph{BSD500} \cite{BSD}. The images are first transferred to $YC_{b}C_{r}$ color space and only utilize the $Y$ channel. The chrominance channels are later simply up-sampled by the bicubic method in the testing process. Then, we generate sub-images at the size of $32\times32$ from images in the dataset with the stride step of 16 pixel. Down-sampling method in \cite{NCSR} is utilized that images are first blurred and then down-sampled with factors of 2, 3 and 4. As a result, around 10 thousand sub-images are obtained for training. The learning rate is initially set as $10^{-4}$ and drops to $10^{-5}$ after 50,000 iterations.

We compare our algorithm with different SR methods including a typical learning-based SR method \cite{NE} (denoted as NE) and two online compensation methods \cite{landmark2013,onlinetmm2016} (respectively denoted as Landmark and GSSR). For fair comparison, we add the retrieved HR reference image to the training set of learning-based method NE. Besides, the intermediate results derived by IHN \cite{IHN} are also shown as the baseline. The baseline is one of the newest deep based SR methods without using external references. The testing images are chosen from the Oxford Building dataset\footnotemark[1] and the online retrieval is also performed over it. There are totally 8 testing images named from (a) to (h) for comparison, as shown in Fig. \ref{fig:f3}. We set $\emph{K}=4$ for the number of reference images. More experimental results can be found on our website\footnotemark[2].
\footnotetext[1]{\url{http://www.robots.ox.ac.uk/~vgg/data/oxbuildings/}}
\footnotetext[2]{\url{http://www.icst.pku.edu.cn/struct/Projects/DualSR.html}}

\renewcommand\arraystretch{1.3}
\begin{table}[t]
\scriptsize
\centering
\caption{\upshape PSNR and SSIM values of different methods. $(\cdot)$ denotes performance gain of the proposed method compared with other methods.}
\label{t1}
	\begin{tabular}{c|c|c|c|c|c|c}
		\hline
		Scale &  Metrics   & NE & Landmark & GSSR  & Baseline & Proposed \\
		\hline
		\multirow{4}[0]{*}{2} & \multirow{2}[0]{*}{PSNR} & 28.40  & 30.41  & 31.39  & 32.56  & \textbf{33.66} \\
		&       & (5.26)  & (3.25)  & (2.27)  & (1.10)  & - \\ \cline{2-7}
		 &    \multirow{2}[0]{*}{SSIM}   & 0.822  & 0.860  & 0.894  & 0.922  & \textbf{0.937}  \\
		&       & (.115)  & (.077)  & (.043)  & (.015)  & - \\
		\hline
		 \multirow{4}[0]{*}{3}& \multirow{2}[0]{*}{PSNR} & 27.25  & 29.31  & 29.20  & 29.48  & \textbf{30.93} \\
		&       & (3.69)  & (1.63)  & (1.74)  & (1.45)  & - \\		\cline{2-7}
        &  \multirow{2}[0]{*}{SSIM}     & 0.796  & 0.826  & 0.840  & 0.849  & \textbf{0.884}  \\
		&       & (.088)  & (.058)  & (.044)  & (.035)  & - \\
		\hline
		\multirow{4}[0]{*}{4} & \multirow{2}[0]{*}{PSNR} & 25.61  & 27.71  & 27.69  & 27.85  & \textbf{29.35} \\
		&       & (3.74)  & (1.64)  & (1.67)  & (1.50)  & - \\		\cline{2-7}
        &    \multirow{2}[0]{*}{SSIM}   & 0.740  & 0.786  & 0.785  & 0.791  & \textbf{0.835}  \\
		&       & (.095)  & (.049)  & (.051)  & (.044)  & - \\
		\hline
	\end{tabular}%
\vspace{-2mm}
\end{table}

\subsection{Experimental Results and Analysis}
\label{sec:experimentsetres}
Table \ref{t1} shows objective results of $8$ chosen images. Our proposed method obtains the best average PSNR and SSIM values in all cases.

Subjective results are shown in Fig. \ref{fig:f4}. The edge-preserving based method NE successfully obtains more sharp edge but fails to reconstruct other more detailed HF signals. Although Landmark has successfully combined some HF signals of HR references, artifacts sometimes are brought by incorrect patch matching results or inappropriate patch blending. Sparse-based method GSSR did not consider position feature of the reference patches. While there are many similar reference patches, more noise are brought into GSSR's SR reults. Edge feature combined baseline method \cite{IHN} has also well reconstructed some HF signal. However, without information from HR references, it fails to reconstruct the detail in complex regions. On the contrary, our method achieves the best result in HF information reconstruction.

\vspace{-4mm}
\begin{table}[h]
\centering
\caption{\upshape PSNR and SSIM values of VDSR and the proposed method.}
\label{t2}
\begin{tabular}{c|c|c|c|c|c|c}
\hline
\multirow{2}{*}{Metrics} & \multicolumn{3}{c|}{VDSR} & \multicolumn{3}{c}{Proposed Method} \\ \cline{2-7}
                         & 2       & 3      & 4      & 2      & 3      & 4     \\ \hline
\multirow{2}{*}{PSNR}    & 33.12   & 29.90  & 28.34  & \textbf{33.94} & \textbf{31.28} & \textbf{30.04} \\ \cline{2-7}
                          & \textbf{0.81}  & \textbf{1.38}  & \textbf{1.70}   & -      & -      & -     \\ \hline
\multirow{2}{*}{SSIM}    & 0.931   & 0.860  & 0.803  & \textbf{0.942} & \textbf{0.892} & \textbf{0.856} \\ \cline{2-7}
                          & \textbf{0.011} & \textbf{0.032} & \textbf{0.053}  & -      & -      & -     \\ \hline
\end{tabular}
\end{table}%
\vspace{-4mm}
We also compare with one of state-of-the-art methods, VDSR\cite{VDSR}. Due to the different bicubic down-sampling configuration, we have retrained our network by utilizing VDSR as the IHN under the new configurat. The results have been shown in Table \ref{t2}. Our method still obtains the gain over VDSR.
\vspace{-3mm}
\section{Conclusion}
\label{sec:conclu}
In this paper, we propose a deep online compensation network for image super-resolution. With the IHF map estimated by IHN, we initially obtain an intermediate SR result by combining the IHF map with a simply up-sampled LR image. Then, the EHF maps are further extracted from online retrieved HR references for compensation. The final SR result is obtained by adding the fused EHF map to the intermediate SR result. Extensive experimental results demonstrate that the proposed method can robustly extract external HF maps from the reference images and significantly improve the SR results based on the compensation brought by the EHF maps.

\bibliographystyle{IEEEbib}
\bibliography{bib_iscas_2018}

\begin{thebibliography}{10}

\bibitem{sun_gradient_2011}
J.~Sun, J.~Sun, Z.~Xu, and H.~Y. Shum,
\newblock ``Gradient profile prior and its applications in image
  super-resolution and enhancement,''
\newblock {\em {IEEE} Transactions on Image Processing}, vol. 20, no. 6, pp.
  1529--1542, 2011.

\bibitem{Breg2}
A.~Marquina and SJ. Osher,
\newblock ``Image super-resolution by {TV}-regularization and bregman
  iteration,''
\newblock {\em Journal of Scientific Computing}, vol. 37, no. 3, pp. 367--382,
  2008.

\bibitem{SCSR}
J.~Yang, J.~Wright, T.~Huang, and Y.~Ma,
\newblock ``Image super-resolution via sparse representation,''
\newblock {\em {IEEE} Transactions on Image Processing}, vol. 19, no. 11, pp.
  2861--2873, 2010.

\bibitem{aplus}
R.~Timofte, V.~De Smet, and L.~Van Gool,
\newblock ``A+: Adjusted anchored neighborhood regression for fast
  super-resolution,''
\newblock in {\em Proc. Asian Conference on Computer Vision}, 2014.

\bibitem{NE}
Y.~Li, J.~Liu, W.~Yang, and Z.~Guo,
\newblock ``Neighborhood regression for edge-preserving image
  super-resolution,''
\newblock in {\em Proc.~IEEE Int'l Conf.~Acoustics, Speech, and Signal
  Processing}, 2015.

\bibitem{SRCNN}
C.~Dong, C.~Chen, K.~He, and X.~Tang,
\newblock ``Learning a deep convolutional network for image super-resolution,''
\newblock in {\em Proc.~European Conference on Computer Vision}, 2014.

\bibitem{DSP}
D.~Liu, Z.~Wang, B.~Wen, J.~Yang, W.~Han, and T.~S. Huang,
\newblock ``Robust single image super-resolution via deep networks with sparse
  prior,''
\newblock {\em {IEEE} Transactions on Image Processing}, vol. 25, no. 7, pp.
  3194--3207, 2016.

\bibitem{VDSR}
J.~Kim, J.~K. Lee, and K.~M. Lee,
\newblock ``Accurate image super-resolution using very deep convolutional
  networks,''
\newblock in {\em Proc.~IEEE Int'l Conf.~Computer Vision and Pattern
  Recognition}, 2016.

\bibitem{IHN}
W.~Yang, J.~Feng, J.~Yang, F.~Zhao, J.~Liu, Z.~Guo, and S.~Yan,
\newblock ``Deep edge guided recurrent residual learning for image
  super-resolution,''
\newblock {\em {IEEE} Transactions on Image Processing}, vol. 26, no. 12, pp.
  5895 -- 5907, 2017.

\bibitem{CVIUVDSR}
W.~Yang, J.~Feng, G.~Xie, J.~Liu, Z.~Guo, and S.~Yan,
\newblock ``Video super-resolution based on spatial-temporal recurrent residual
  networks,''
\newblock {\em Computer Vision and Image Understanding}, 2017.

\bibitem{timofte2016semantic}
R.~Timofte, VD. Smet, and LV. Gool,
\newblock ``Semantic super-resolution: When and where is it useful?,''
\newblock {\em Computer Vision and Image Understanding}, 2016.

\bibitem{landmark2013}
H.~Yue, X.~Sun, J.~Yang, and F.~Wu,
\newblock ``Landmark image super-resolution by retrieving web images,''
\newblock {\em {IEEE} Transactions on Image Processing}, vol. 22, no. 12, pp.
  4865--4875, 2013.

\bibitem{SRreference}
Y.~Li, W.~Dong, G.~Shi, and X.~Xie,
\newblock ``Learning parametric distributions for image super-resolution: Where
  patch matching meets sparse coding,''
\newblock in {\em Proc.~IEEE Int'l Conf.~Computer Vision}, 2015.

\bibitem{onlinetmm2016}
J.~Liu, W.~Yang, X.~Zhang, and Z.~Guo,
\newblock ``Retrieval compensated group structured sparsity for image
  super-resolution,''
\newblock {\em {IEEE} Transactions on Multimedia}, vol. 19, no. 2, pp.
  302--316, 2017.

\bibitem{BSD}
P.~Arbelaez, M.~Maire, C.~Fowlkes, and J.~Malik,
\newblock ``Contour detection and hierarchical image segmentation,''
\newblock {\em {IEEE} Transactions on Pattern Analysis and Machine
  Intelligence}, vol. 33, no. 5, pp. 898--916, 2011.

\bibitem{NCSR}
Y.~Li, W.~Dong, G.~Shi, and X.~Xie,
\newblock ``Learning parametric distributions for image super-resolution: Where
  patch matching meets sparse coding view document,''
\newblock in {\em Proc.~IEEE Int'l Conf.~Computer Vision}, 2015.

\end{thebibliography}
\end{document}